\documentclass[pdflatex,sn-mathphys-num]{sn-jnl}

\usepackage{graphicx}%
\usepackage{multirow}%
\usepackage{amsmath,amssymb,amsfonts}%
\usepackage{amsthm}%
\usepackage{mathrsfs}%
\usepackage[title]{appendix}%
\usepackage{xcolor}%
\usepackage{textcomp}%
\usepackage{manyfoot}%
\usepackage{booktabs}%
\usepackage{algorithm}%
\usepackage{algorithmicx}%
\usepackage{algpseudocode}%
\usepackage{listings}%

\usepackage{subcaption} 
\usepackage{minted}
\usepackage{tcolorbox}
\usepackage{url}
\usepackage{pdflscape}

\usepackage{hyperref}

\let\orcidlogo\relax  
\usepackage{orcidlink}

\theoremstyle{thmstyleone}%
\theoremstyle{thmstyletwo}%

\theoremstyle{thmstylethree}%

\begin{document}

\title[Article Title]{Construction of a Battery Research Knowledge Graph using a Global Open Catalog}


\author[1]{\fnm{Luca} \sur{Foppiano}\orcidlink{0000-0002-6114-6164}}\email{luca@sciencialab.com}

\author*[2,3]{\fnm{Sae} \sur{Dieb}\orcidlink{0000-0002-8111-2009}}\email{dieb.sae@nims.co.jp}

\author[4]{\fnm{Malik} \sur{Zain}}\email{zain.malik@unf.edu}

\author[3]{\fnm{Kazuki} \sur{Kasama}}\email{kazuki@igroupjapan.com}

\author[2]{\fnm{Keitaro} \sur{Sodeyama}\orcidlink{0000-0002-9228-0729}}\email{keitaro.sodeyama@nims.go.jp}

\author[3,5]{\fnm{Mikiko} \sur{Tanifuji}\orcidlink{0000-0001-5284-6364}}\email{tanifuji@nii.ac.jp}

\affil[1]{\orgname{ScienciaLAB}, \orgaddress{\city{Loulé}, \country{Portugal}}}

\affil[2]{\orgdiv{CBRM}, \orgname{National Institute for Materials Science}, \orgaddress{\city{Tsukuba}, \country{Japan}}}

\affil[3]{\orgname{CyberMate LLC}, \orgaddress{\city{Tokyo}, \country{Japan}}}

\affil[4]{\orgdiv{School of Computing}, \orgname{University of North Florida}, \orgaddress{\city{Jacksonville}, \country{United States}}}

\affil[5]{\orgname{National Institute of Informatics}, \orgaddress{\city{Tokyo}, \country{Japan}}}


\abstract{Battery research is a rapidly growing and highly interdisciplinary field, making it increasingly difficult to track relevant expertise and identify potential collaborators across institutional boundaries. In this work, we present a pipeline for constructing an author-centric knowledge graph of battery research built on OpenAlex, a large-scale open bibliographic catalogue. For each author, we derive a weighted research descriptors vector that combines coarse-grained OpenAlex concepts with fine-grained keyphrases extracted from titles and abstracts using KeyBERT with ChatGPT (gpt-3.5-turbo) as the backend model, selected after evaluating multiple alternatives. Vector components are weighted by research descriptor origin, authorship position, and temporal recency. The framework is applied to a corpus of 189,581 battery-related works. The resulting vectors support author–author similarity computation, community detection, and exploratory search through a browser-based interface. The knowledge graph is then serialized in RDF and linked to Wikidata identifiers, making it interoperable with external linked open data sources and extensible beyond the battery domain. Unlike prior author-centric analyses confined to institutional repositories, our approach operates at cross-institutional scale and grounds similarity in domain semantics rather than citation or co-authorship structure alone.}




\maketitle

\begin{center}
  \fbox{\parbox{0.85\linewidth}{\centering\small
    \textbf{Preprint — under peer review.} This manuscript has not been 
    formally accepted for publication.
  }}
\end{center}

\section{Introduction}
Batteries are a fundamental technology enabling modern energy systems. Battery research intersects with materials science, electrochemistry, safety, manufacturing, and other research areas. With the increasing demand for a green transition toward renewable energy technologies, a growing number of highly interdisciplinary scientific papers are being published~\cite{ma2021the}. 
Given this information overload and the fragmentation of battery knowledge across multiple research fields, it is difficult to track relevant concepts and terminology. There is a need for a structured representation of concepts related to battery research that supports easier exploration of the field.

Existing works have been proposed to provide a concept map for battery-specific field~\cite{clark2025semantic, zhang2024materials}. However, to our knowledge, they are organized around documents or entities, and do not support author-level discovery. 
An author-level concept map enables quantitative similarity retrieval, potential collaborations, and expertise identification.

In this paper, we introduce a pipeline to construct a reusable, visual, author-centric knowledge graph of battery research descriptors. 
The graph aggregates battery-relevant descriptors across each author’s publications as a vector representation of the total author research output. 
This structure enables explainable author similarity search (such as collaborator discovery, reviewer recommendation) and community detection grounded in domain semantics rather than solely in citation or co-authorship structure. 

In our approach, we used natural language processing techniques on a large dataset constructed using an open catalogue named OpenAlex~\cite{priem2022openalex}. 
OpenAlex provides a rich bibliographic graph composed of aggregated metadata for works, authors, and organizations. 
Building on OpenAlex offers several advantages: documents are already tagged with general concepts, enabling reproducible data processing; the platform is highly scalable; and its data is continuously updated.
However, OpenAlex lacks fine-grained domain semantics. 
We therefore complemented OpenAlex concepts with keyphrases extracted using KeyBERT~\cite{grootendorst2020keybert} -- selected after benchmarking against SentenceTransformers and domain-tuned BERT variants -- to capture battery-specific terminology at a finer granularity.
We constructed author-level research descriptors using three different weighting factors (Section~\ref{sec:author-vectors}): the descriptor origin (OpenAlex concepts vs. keyphrases), temporal recency (older publication periods were down-weighted to emphasize current expertise), and author position (first-author publications were weighted more heavily as they better reflect primary research focus).
Using the resulting author research descriptor vectors, we computed author–author similarity to identify related researchers. 
A browser-based interface was implemented to visualize the graph with functions related to exploratory search. 
Finally, we serialized the graph into the RDF (Resource Description Framework) format~\cite{manola2004rdf} to support standardized integration with external knowledge graphs, enabling reuse and extension beyond the battery field. 

The remainder of this paper is organized as follows. Section~\ref{sec:related-work} reviews related work, followed by our method in Section~\ref{sec:method}, covering dataset preparation, keyword extraction, data aggregation, and author descriptor construction. Sections~\ref{sec:word-cloud-representation} and~\ref{sec:similarity-map} present the visualization and similarity map, Section~\ref{sec:rdf} describes the RDF serialization, and Section~\ref{sec:conclusions} concludes the paper.

\section{Related Work}
\label{sec:related-work}

Text mining techniques have been widely applied in materials science to extract structured knowledge from large collections of scientific articles, supporting the automatic construction of specialized datasets across a broad range of subdomains~\cite{lfoppiano2023automatic,court_magnetic_2020,kononova2019textminted, huang2020database, dieb2012automatic, dieb2015extraction, dieb2016nadev, foppiano2021supermat}. 

Building on these extraction methods, several general-purpose NLP approaches have been proposed for keyword selection, topic identification, and trend analysis in scientific literature. 
Trend detection has received particular attention, with approaches ranging from research trend prediction over time~\cite{charnine2021visualization, jamali2011article} to the identification of ``bursty research topics'' through co-word analysis and matrix factorization~\cite{katsurai2019trendets}.
Topic modeling methods have been surveyed extensively by \citet{rani2021topic}, who covered classical techniques in materials science and engineering but largely omitted more recent approaches such as Latent Topic Semantic Graphs~\cite{law2018ltsg}. 
\citet{nadim2023comparative} compared several approaches for keyword extraction and topic modeling, finding that BERT-based encoders yield stable results while LLMs achieved stronger performance, though at the cost of sensitivity to prompt design~\cite{chataut2024comparative, jia2025llm, maragheh2023llm}.
A related line of work addresses scholarly network analysis and author-level representation. 
\citet{xu2018anetwork} proposed a multi-network embedding framework for author disambiguation, while \citet{nie2021construction} described a system for tracking research development around a specific subject. 
\citet{mueller2017semantic} demonstrated that title-level semantic representations can be effectively obtained by averaging word embeddings, highlighting the importance of in-domain training data and temporal information for publication disambiguation.
\citet{schafermeier2023research} proposed a graph structure to compute author research ranking in specific domains, and citation graph analysis has also been applied to research lineage identification through citation classification~\cite{ghosal2022towards}.
At the institutional level, \citet{Dieb01012021} conducted author-centric analyses within a single repository, providing valuable insights into local research dynamics but inherently limited in scope and cross-institutional coverage.
Within the materials science domain specifically, a complementary line of work focused on building structured terminological resources from text corpora rather than extracting specific property data. \citet{zhang2024materials} proposed MGED-KG, a materials terminology knowledge graph automatically constructed via NLP from a bilingual Chinese-English dictionary corpus. Beyond cataloging terminology, MGED-KG was embedded into a materials data platform to support query expansion, term recommendation, and data discovery, demonstrating how NLP-derived knowledge graphs can serve as a semantic backbone for cross-domain data integration in materials science. 
The battery subdomain has also seen dedicated investment in semantic infrastructure to support knowledge formalization and data interoperability. \citet{clark2025semantic} surveyed the role of semantic technologies — ontologies, RDF-based vocabularies, and open-source tooling — in structuring battery research knowledge in a machine-readable form, with a focus on resources developed within the BATTERY 2030+ project~\cite{battery2030plus}. 
Their work introduced the Battery Knowledge Base (BKB), a web platform that links domain ontologies to wiki-style articles, datasets, and multimedia content, serving as a centralized hub for knowledge sharing across the battery community. This ontology-driven approach complemented text mining pipelines by providing the controlled vocabularies against which extracted concepts can be normalized and linked.
Work on large-scale scholarly infrastructures such as OpenAlex~\cite{priem2022openalex} has enabled new forms of analysis: \citet{farber2023semopenalex} constructed an RDF-like ontology centered on the citation graph, whereas our work instead focuses on topic and keyword proximity within selected subsets of the OpenAlex corpus.
In contrast with prior work, none of the approaches above combines author-centric analysis, domain-specific research descriptors, and cross-institutional scope within a single framework. 
The present work addresses this gap by constructing an author-level representation of the battery literature — built directly on the OpenAlex corpus — that integrates thematic descriptors, co-authorship networks, and temporal weighting, with possible generalization to other domains.

\section{Method}
\label{sec:method}
\begin{figure}[htbp]
  \centering
  \includegraphics[height=0.8\textheight,keepaspectratio]{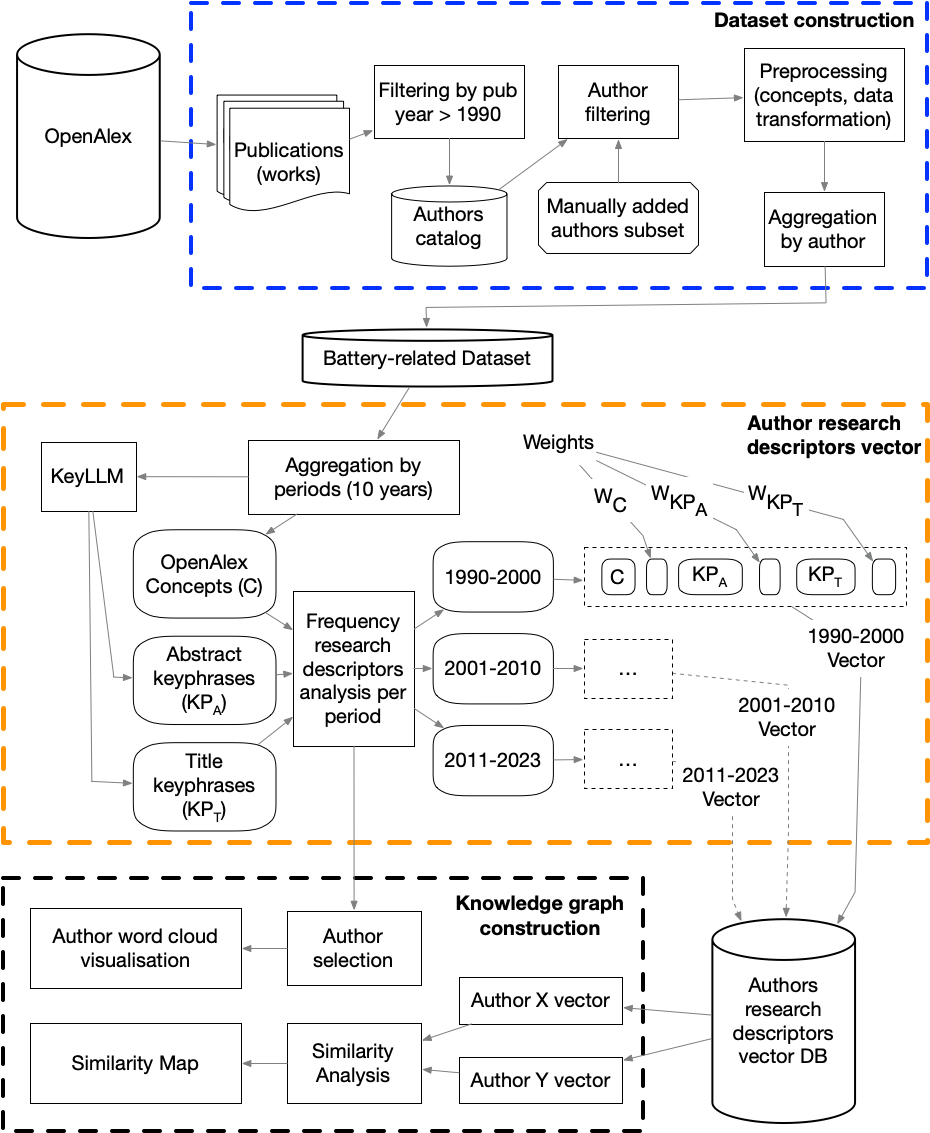}
  \caption{Overview of the proposed pipeline. Dataset construction (Section~\ref{sec:dataset-preparation}) collects, filters, and preprocesses data from OpenAlex, extracting OpenAlex concepts in the process. The resulting battery-related dataset is enriched with keyphrases extracted from titles and abstracts (Section~\ref{sec:keyword-extraction}). Research descriptors are then computed with consideration of descriptor origin, temporal period, and authorship position (Section~\ref{sec:data-aggregation}) to generate a per-author research descriptor vector database (Section~\ref{sec:author-vectors}). Finally, the knowledge graph construction supports the generation of word clouds (Section~\ref{sec:word-cloud-representation}) and the author similarity map (Section~\ref{sec:similarity-map}).}
  \label{fig:overview}
\end{figure}

The outline of the proposed workflow is represented in Figure~\ref{fig:overview}.
First, we collected battery-related scholarly publications from OpenAlex and preprocessed them into a clean, standardized dataset, filtering out irrelevant records and extracting OpenAlex concepts in the process.
We then performed keyphrase extraction from titles and abstracts, and combined these with the OpenAlex concepts to build research descriptors for each publication.
These descriptors were aggregated at the author level, weighted by origin (concept/keyphrase), temporal period and authorship position, to produce an author-level vector database carrying information about publications, co-authorship, and thematic focus over time.
Finally, author research descriptor vectors were used to generate a similarity map revealing author relations through shared research descriptors.

\subsection{Dataset preparation}
\label{sec:dataset-preparation}

The dataset used for this study was a preprocessed subset from the OpenAlex catalogue.
OpenAlex is a fully open bibliographic database of scholarly works, authors, institutions, venues, and concepts, launched in 2022 as a successor to Microsoft Academic Graph (MAG)~\cite{sinha2015anOverview}. It indexes over 250 million scholarly works drawn from sources such as Crossref, PubMed, institutional repositories, and open-access archives, making it one of the largest freely available academic knowledge bases. 
The data is structured around five core entity types — Works, Authors, Sources, Institutions, and Concepts — linked through typed relationships such as authorship, citation, affiliation, and topical classification. Each entity is assigned a persistent OpenAlex identifier and enriched with metadata including DOIs, ORCIDs, open-access status, citation counts, and disambiguated author profiles. 
The concept taxonomy spans six hierarchical levels — from broad domains such as ``Physics'' down to fine-grained sub-topics. Each work is associated with multiple concepts derived from its title, abstract, and host venue title, using an automated classifier trained on the MAG corpus; a confidence score between 0 and 1 accompanies each assigned concept — though their ancestor concepts are propagated automatically up the hierarchy regardless of score, meaning a work tagged with ``Battery (electricity)'' (C555008776, level 3) will also carry its ancestors: level-0 (``Physics''), level-1 (``Quantum Mechanics'', and ``Thermodynamic''), and level-2 (``Power (physics)'').

In this work, we first retrieved the Works from the OpenAlex API by querying records associated with the identifier ``C555008776'' (``Battery (electricity)'') and we collected 189,581 records that corresponded to 356,103 authors. 
We then filtered publications older than 1990 or without publication date to reduce the noise brought by older and imprecise publication data. 
As a case demonstration, we refined the collected dataset to include publications from the top 10,000 authors, sorted by number of publications. This was to provide a proof of concept with reduced computational requirements.
We indexed authors using a unique key that was generated using a normalized OpenAlex ID. 
In addition, we added a subset of researchers from our institution that are specialized in battery research for validation purposes.

The dataset preparation included those post-processing steps: 
\begin{itemize}
    \item we pruned the concept list of the generic ``battery'' concept, which is shared across all collected publications and does not provide any additional granular information;
    \item we removed all concepts with a zero confidence score, as these are propagated automatically for hierarchical navigation purposes and introduce noise into our representations; and
    \item we applied data transformation by selecting a subset of fields (identifiers, title, publication metadata, authors and affiliations, abstract, and associated concepts with scores).
\end{itemize}

It is worth noting that, as an aggregated resource, OpenAlex is not free from noise: some records carry incomplete or erroneous metadata, such as missing affiliation information. Correcting such inconsistencies was outside the scope of this work.

\subsection{Keyphrases extraction}
\label{sec:keyword-extraction}

OpenAlex concepts are too coarse for our purposes: spanning all scientific disciplines, they would produce overly broad author similarity matches within a single research domain.
For this reason, we complemented the OpenAlex concepts with more domain-specific text-based keyphrases, extracted from the titles and abstracts of the articles. 
We used KeyBERT~\cite{grootendorst2020keybert}, an open-source library based on BERT-embeddings~\cite{devlin2019bert} and cosine similarity that, given a text in input, finds the keyphrases that are the most representative for the given text. 
We selected among four models: the original SentenceTransformers~\cite{reimers2019sentencebert}, two battery-domain-pretrained BERT variants~\cite{huang2022batterybert} and a Large Language Model (LLM), OpenAI ChatGPT (gpt-3.5-turbo). 

To select the best model, we constructed an evaluation framework composed of a sample of 100 articles and a scoring algorithm. 
The ground truth was extracted using Grobid~\cite{grobid} from the 100 PDF articles as a silver-standard reference, while acknowledging that keyword metadata is incomplete and heterogeneous across venues. 
The scoring algorithm was designed to capture broader semantic overlap between two lists of heterogeneous elements. 
Given $P_d$ is the keyphrases predicted from the models, and $E_d$ the keyphrases expected for document $d$. 
Let $k_d = \min(|P_d|, |E_d|)$, and let $\hat{P}_d \subset P_d$ denote the $k_d$ highest-confidence predicted keyphrases, with $\hat{E}_d \subset E_d$ the corresponding trimmed expected set, $|\hat{E}_d| = k_d$.
Define the threshold ($\tau \in [0, 1]$) cosine similarity of a predicted keyphrase $p$ against an expected keyword $e$ as:
\begin{equation}
    \phi(e, p) = \max\!\bigl(0,\, \cos(e, p) \cdot \mathbb{I}[\cos(e, p) > \tau]\bigr)
\end{equation}

The per-expected-keyphrases score aggregated over predictions is:
\begin{equation}
    s_e^{(d)} = \frac{1}{|\hat{P}_d|} \sum_{p \,\in\, \hat{P}_d} \phi(e, p)
\end{equation}

The per-document score is:
\begin{equation}
    s_d = \frac{1}{|\hat{E}_d|} \sum_{e \,\in\, \hat{E}_d} s_e^{(d)}
    = \frac{1}{k_d^2} \sum_{e \,\in\, \hat{E}_d}\ \sum_{p \,\in\, \hat{P}_d} \phi(e, p)
\end{equation}

And the final dataset-level similarity score as an average over ${N}$ documents:
\begin{equation}
    S = \frac{1}{N} \sum_{d=1}^{N} s_d
    = \frac{1}{N} \sum_{d=1}^{N} \frac{1}{k_d^2} \sum_{e \,\in\, \hat{E}_d}\ \sum_{p \,\in\, \hat{P}_d} \phi(e, p)
\end{equation}

The evaluation scores in Table~\ref{tab:avg_similarity} indicate each model's achieved dataset-level score $S$. The OpenAI ChatGPT LLM achieved the highest score with Sentence-Transformer and domain-tuned BERT models close behind. 

\begin{table}[htp]
\centering
\begin{tabular}{l c}
\hline
\textbf{Method} & \textbf{Avg. Similarity} \\
\hline
ChatGPT (chatgpt-3.5-turbo) & 0.6781 \\
BatterySciBERT\_cased & 0.6698 \\
BatteryOnlyBERT & 0.6677 \\
SentenceTransformers & 0.6665 \\
BatterySciBERT\_uncased & 0.5423 \\
\hline
\end{tabular}
\caption{Average dataset-level similarity score $S$ for keyphrases extracted by different models, evaluated against article-supplied keywords as ground truth, on a sample of 100 documents.}
\label{tab:avg_similarity}
\end{table}

After selecting the LLM-based method for keyphrase extraction, we processed the dataset to extract keyphrases from titles and abstracts. We limited the extracted keyphrases, sorted by similarity, to a maximum of two for title keyphrases ${KP}_T$, and 10 abstract keyphrases ${KP}_A$ (Figure~\ref{fig:overview}). 

\subsection{Data aggregation}
\label{sec:data-aggregation}

Once keyphrase extraction was completed, the data was aggregated at the author level.
For each author, we collected their publications and computed the frequency of each research descriptor across their entire publication record. For each temporal period, we recorded the total publication count broken down by authorship role — corresponding, first, and non-first author — as well as the number of publications shared with each co-author. 
The aggregated data was organized as illustrated in Listing~\ref{lst:author-data-structure}.

\lstset{texcl=true,basicstyle=\small\sf,commentstyle=\small\rm,mathescape=true,escapeinside={(*}{*)}}

\begin{lstlisting}[
  caption={Example of the authors aggregated data structure},
  label={lst:author-data-structure},
  frame=single,
  basicstyle=\small\ttfamily,
  breaklines=true,
  rulecolor=\color{black},
  abovecaptionskip=0pt
]
{
  "author1": {
    "1990-2000": { ... },
    "2001-2010": { ... },
    "2011-2023": {
      "nb_publications": 200,
      "nb_publications_first_author": 10,
      "nb_publications_non_first_author": 190,
      "non_first_author": {
        "concepts": {
          "concept1": { "freq": 123, "avg_confidence_score": 0.8 }
        },
        "keyphrases": {
          "keyphrase1": { "freq": 123, "avg_confidence_score": 0.2 }
        },
        "co_authors": {
          "co_author_id1": 1,
          "co_author_id2": 33
        }
      },
      "first_author": {
        "concepts": {
          "concept1": { "freq": 123, "avg_confidence_score": 0.8 }
        },
        "keyphrases": {
          "keyphrase1": { "freq": 123, "avg_confidence_score": 0.8 }
        },
        "co_authors": {
          "co_author_id1": 20,
          "co_author_id2": 124,
          "co_author_id3": 22
        }
      }
    }
  }
}
\end{lstlisting}

\subsection{Author research descriptors vector generation}
\label{sec:author-vectors}


We constructed a weighted research descriptors-frequency vector $\vec{v}_a \in \mathbb{R}^{|\mathcal{V}|}_{\geq 0}$ for each author $a$ based on their publication history. 
The vocabulary $\mathcal{V}$ (a set of research descriptors) was built by selecting the top-1000 most frequent research descriptors collected so far — from OpenAlex concepts and keyphrases — then merging and updating frequencies for overlapping entries.
All research descriptors were normalized via lowercasing and singularization over the vocabulary $\mathcal{V}$.

We processed each author's publications across three temporal periods $\{p_0, p_1, p_2\}$ corresponding to 1990--2000, 2001--2010, and 2011--2023 respectively, applying decay factor $\mathbf{f}_{pj}$ to each period, where older periods received a higher value to progressively suppress their contribution and emphasize recent work.
The final author vector was given by the following equation:
\begin{equation}
\label{eq:author_vector}
    \vec{v}_a \;=\; \sum_{j=0}^{2} \mathbf{f}_{pj} \cdot \vec{v}_{p_j}
\end{equation}
This equation accounted for temporal recency of the author's publications. 
Where $\vec{v}_{pj}$ was the temporal period vector that was constructed from two vectors that considered the author position (first author, non-first author): 

\begin{equation}
    \vec{v}_{pj} \;=\; w_{first} \cdot \vec{D}_{pj}
               \;+\; w_{nonfirst} \cdot \vec{D'}_{pj}
\end{equation}

where $\vec{D}_{pj}$ and $\vec{D'}_{pj}$ were vectors computed based on descriptors where the researcher was the first author and non-first author in the publication, respectively. 
Higher weight was given to research descriptors vectors of first-author publications. 

Finally, 
\begin{equation}
\vec{D}_{pj} = w_c \cdot \vec{C} + w_{pa} \cdot \vec{K}_{PA} + w_{pt} \cdot \vec{K}_{PT}
\end{equation}

where $\vec{C}$ was composed of the author-relative importance values of each concept, where each element was given by the following equation:

\begin{equation}
    \vec{C}\,[\operatorname{i}] \; \;=\; \operatorname{freq}(c_i) \cdot \operatorname{score}(c_i)
\end{equation}

where $\operatorname{freq}(c_i)$ and $\operatorname{score}(c_i)$ were the frequency and the OpenAlex relevance score, respectively, of the i-th concept in the vocabulary $\mathcal{V}$.

In a similar manner, the vectors $\vec{K}_{PA}$, and $\vec{K}_{PT}$ were calculated, without considering the OpenAlex relevance score as the keyphrases were not extracted from OpenAlex.

\subsection{Word cloud representation}
\label{sec:word-cloud-representation}

\begin{figure}[htbp]
  \centering

  \begin{subfigure}[b]{0.49\linewidth}
    \centering
    \includegraphics[width=\linewidth]{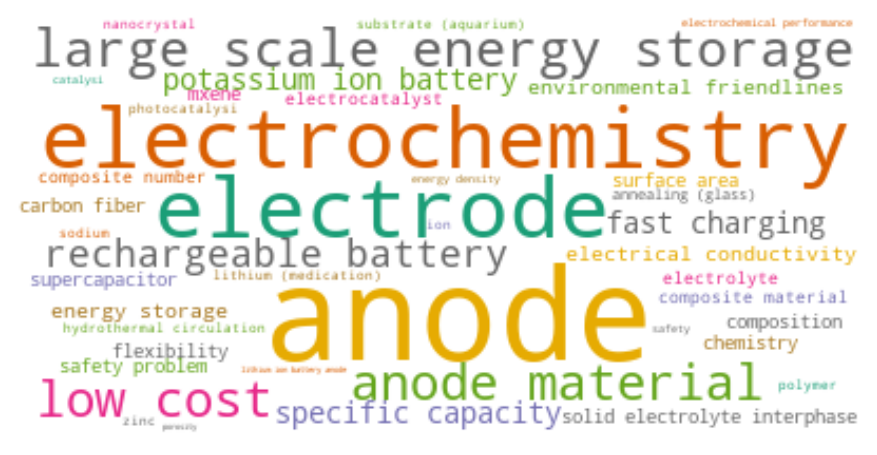}
    \caption{Author 1}
    \label{fig:wordcloud-a5002212606}
  \end{subfigure}\hfill
  \begin{subfigure}[b]{0.49\linewidth}
    \centering
    \includegraphics[width=\linewidth]{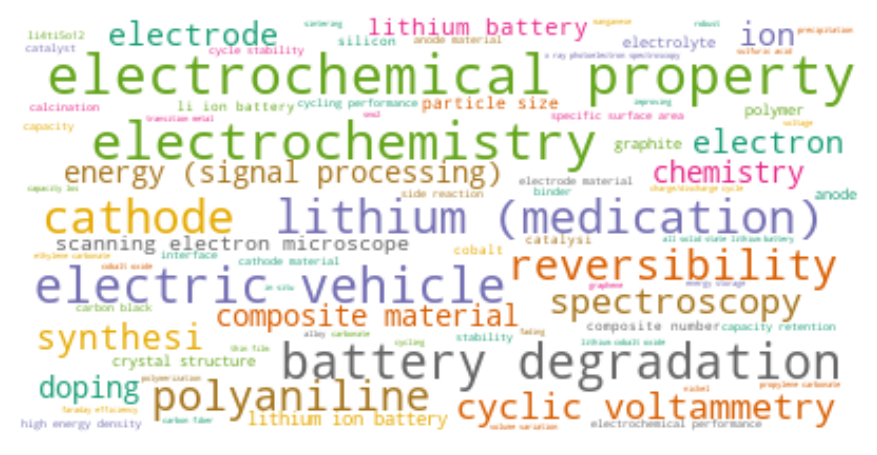}
    \caption{Author 2}
    \label{fig:wordcloud-a5010651513}
  \end{subfigure}

  \caption{Examples of generated word clouds representation of author research descriptors vectors for two authors in battery research.}
  \label{fig:wordcloud-examples}
\end{figure}

We generated word cloud visualizations to provide an interpretable representation of each author's research profile. 
For each author, we constructed a frequency dictionary by mapping research descriptors to their respective vector values, which encoded the relative importance of each research descriptor in the author's publication history. 
We then generated word clouds using the WordCloud Python library~\cite{mueller2023wordcloud} with the font size of each research descriptor proportional to its relevance score in the author vector, allowing dominant research themes to be immediately recognizable. We saved each visualization as a PNG file named after the author's OpenAlex identifier, producing a complete set of visual research profiles for the entire dataset. Figure~\ref{fig:wordcloud-examples} provides two examples for two researchers whose main research is about electrochemistry, with the first author focusing on energy storage, and the second author on lithium-ion batteries (names and affiliations of those authors have been omitted to preserve anonymity).

\subsection{Similarity maps}
\label{sec:similarity-map}

The similarity map is an interactive visualization showing the author graphs through the research descriptors. 

\begin{figure}[htbp]
  \centering

  \begin{subfigure}[t]{0.48\linewidth}
    \centering
    \includegraphics[width=\linewidth]{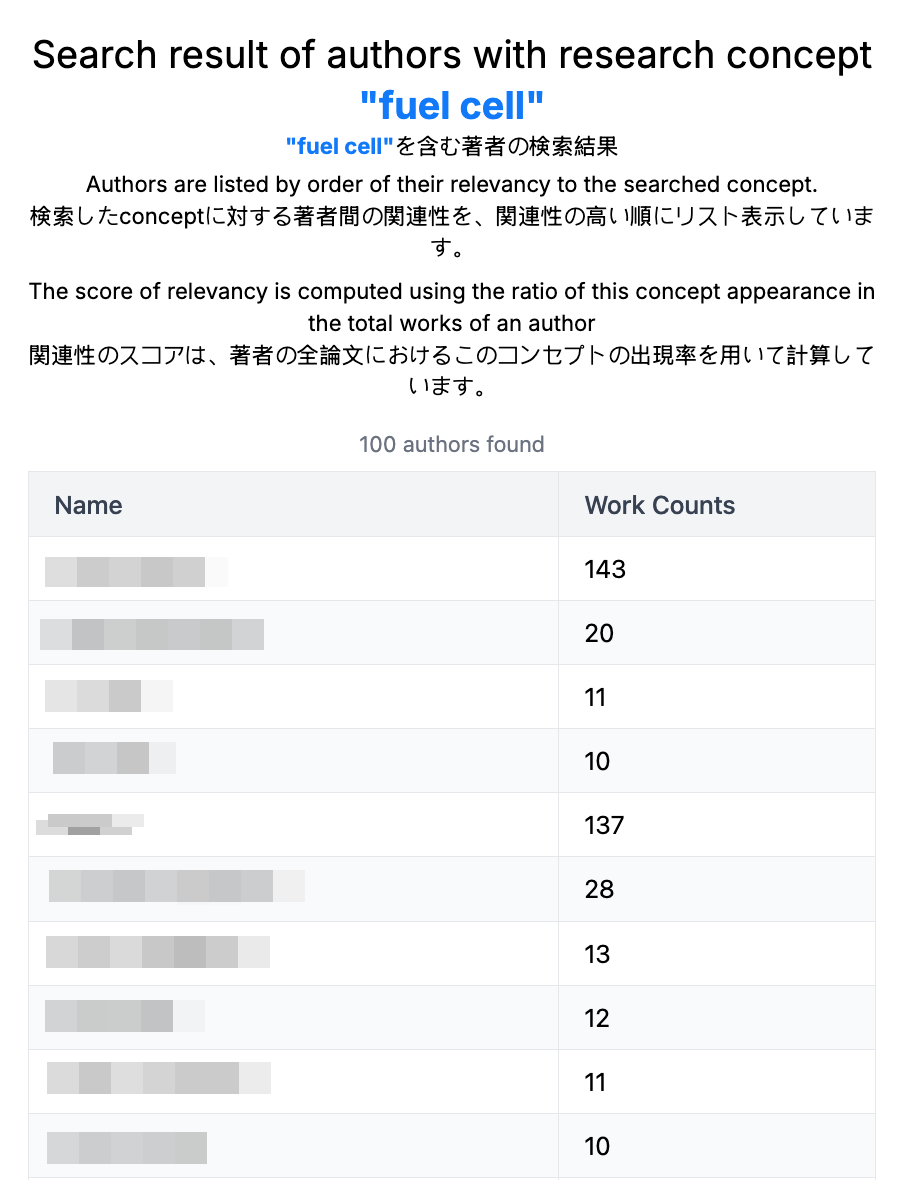}
    \caption{List of authors by concept (research descriptor).}
    \label{fig:search-by-concept}
  \end{subfigure}\hfill
  \begin{subfigure}[t]{0.48\linewidth}
    \centering
    \includegraphics[width=\linewidth]{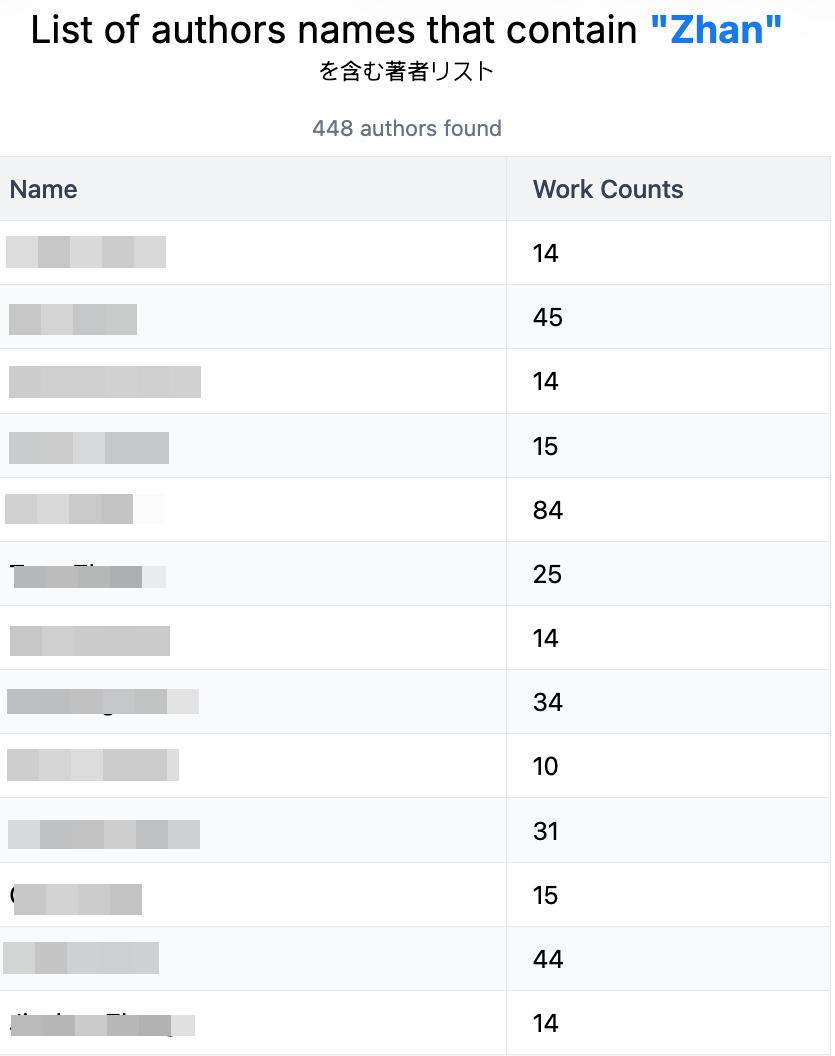}
    \caption{Author search results.}
    \label{fig:search-by-author}
  \end{subfigure}

  \caption{Search results by author name and by concept (research descriptor). Author names have been omitted to preserve anonymity.}
  \label{fig:Result search by author name and concept}
\end{figure}

The map is author-centric: the researcher can be selected by a list populated by either a search by author (Figure~\ref{fig:search-by-author}), or by concept (here we refer to the research descriptor) (Figure~\ref{fig:search-by-concept}). 

\begin{figure}[htbp]
  \centering
  \begin{subfigure}[t]{0.67\linewidth}
    \centering
    \includegraphics[width=\linewidth]{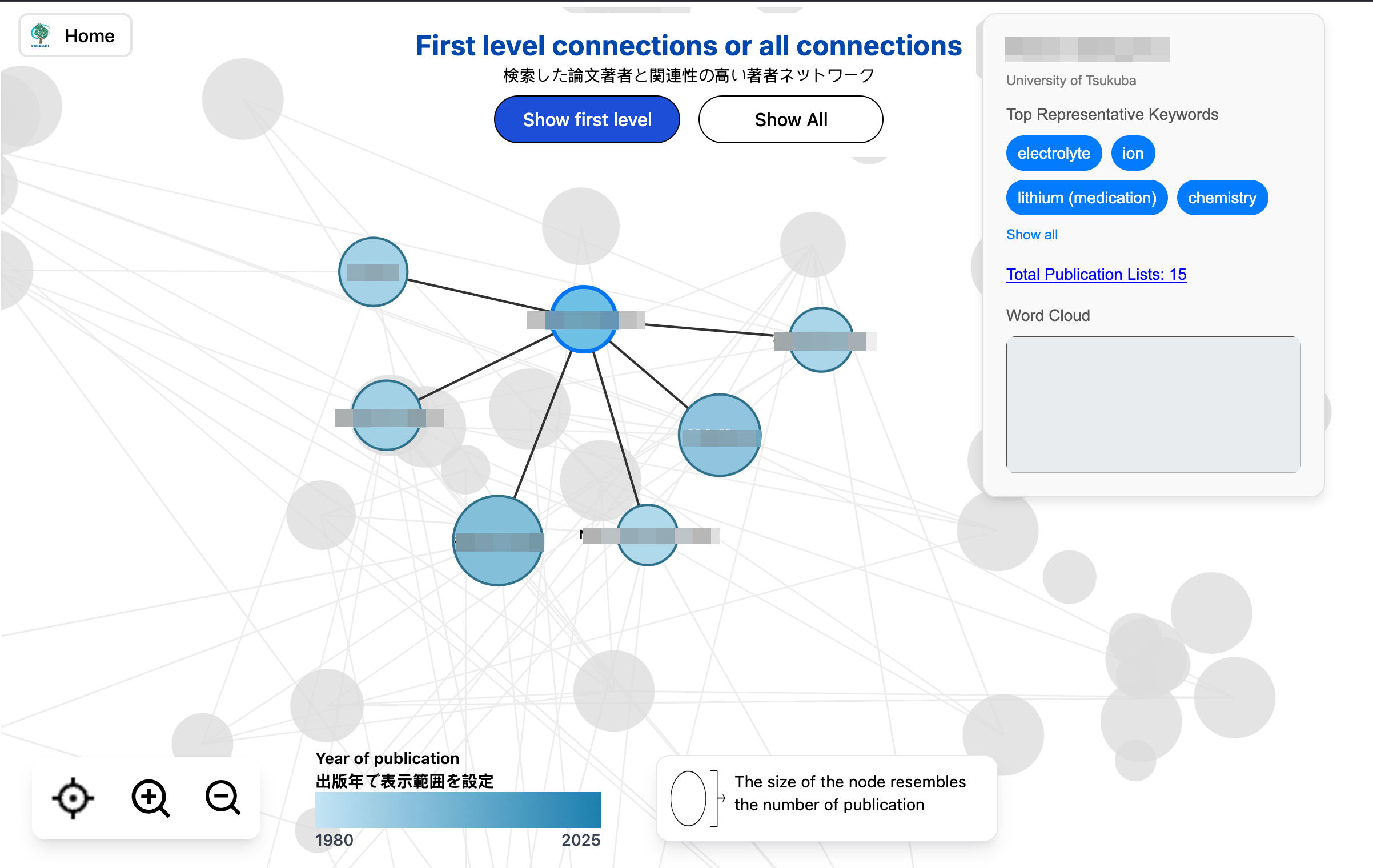}
    \caption{Map of a battery-related author's first-level connections.}
    \label{fig:similarity-map-first-level}
  \end{subfigure}\hfill
  \begin{subfigure}[t]{0.28\linewidth}
    \centering
    \includegraphics[width=\linewidth]{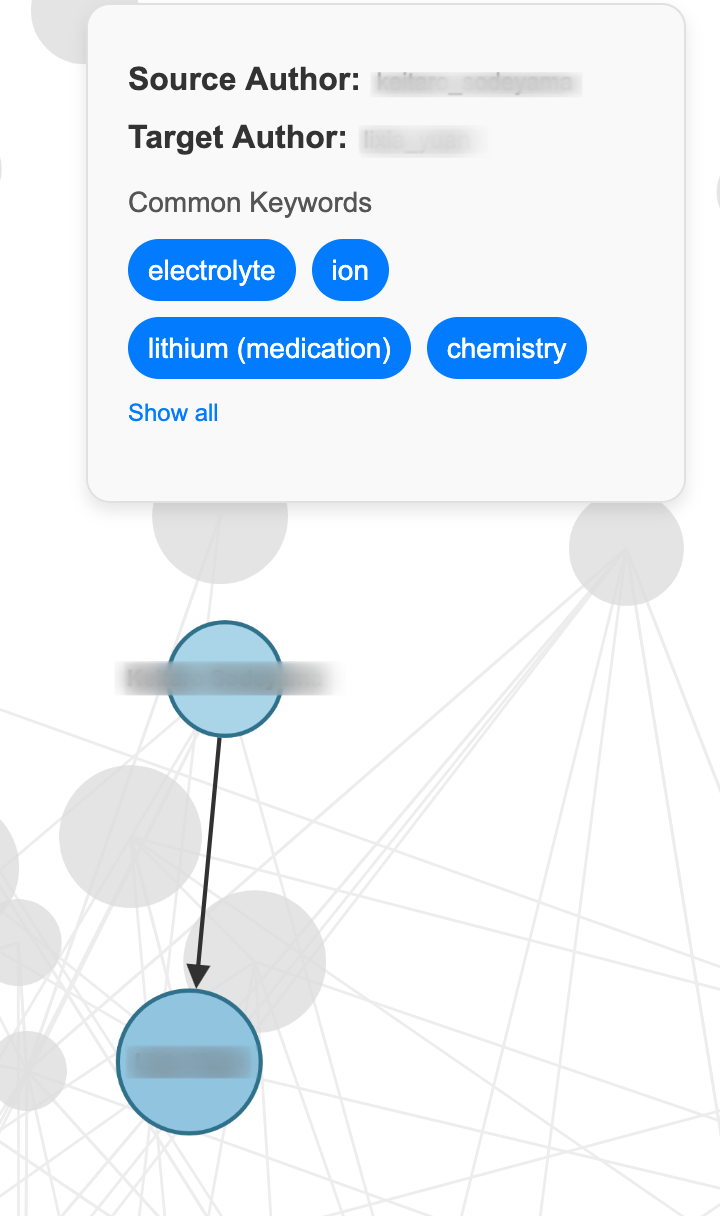}
    \caption{Research descriptors similarity.}
    \label{fig:arc-examination}
  \end{subfigure}

  \caption{Example of similarity map of the first-level connections of a battery-related author (Figure~\ref{fig:similarity-map-first-level}) and the shared research descriptors behind each connection (Figure ~\ref{fig:arc-examination}). In this figure specifically, "the representative keywords" refer to the research descriptors. Author names have been omitted to preserve anonymity.}
  \label{fig:similarity-map-views}
\end{figure}

Selecting the author from the list opens the knowledge graph representation, where the similarity map centered on that author is displayed (Figure~\ref{fig:similarity-map-first-level}). 
Each author is represented by a circle that indicates the number of publications. 
When an author is selected, the primary connections are visualized in blue (authors that shared research descriptors) and in yellow the connections of higher ranks (authors connected indirectly).

In this section, we use information from one of the authors of this work, who has contributed to the field of battery research~\cite{liu2023distinct, tomoaki2024ether, sodeyama2014sacrificial, sodeyama2018liquid}.  
Figure~\ref{fig:similarity-map-first-level} shows the first level connections in blue, indicating other battery research-related authors and one of the co-authors. These connections could be used to foster future collaboration with the authors in the similarity map.

When clicking on an arc, the interface shows the common research descriptors between the two authors, illustrating possible overlap that occurred in the past (Figure~\ref{fig:arc-examination}).
Those elements can be further examined to find other authors that share the same research descriptors ranked by publication count (Figure~\ref{fig:arc-examination}). 

\begin{figure}[htbp]
  \centering
  \includegraphics[width=\linewidth]{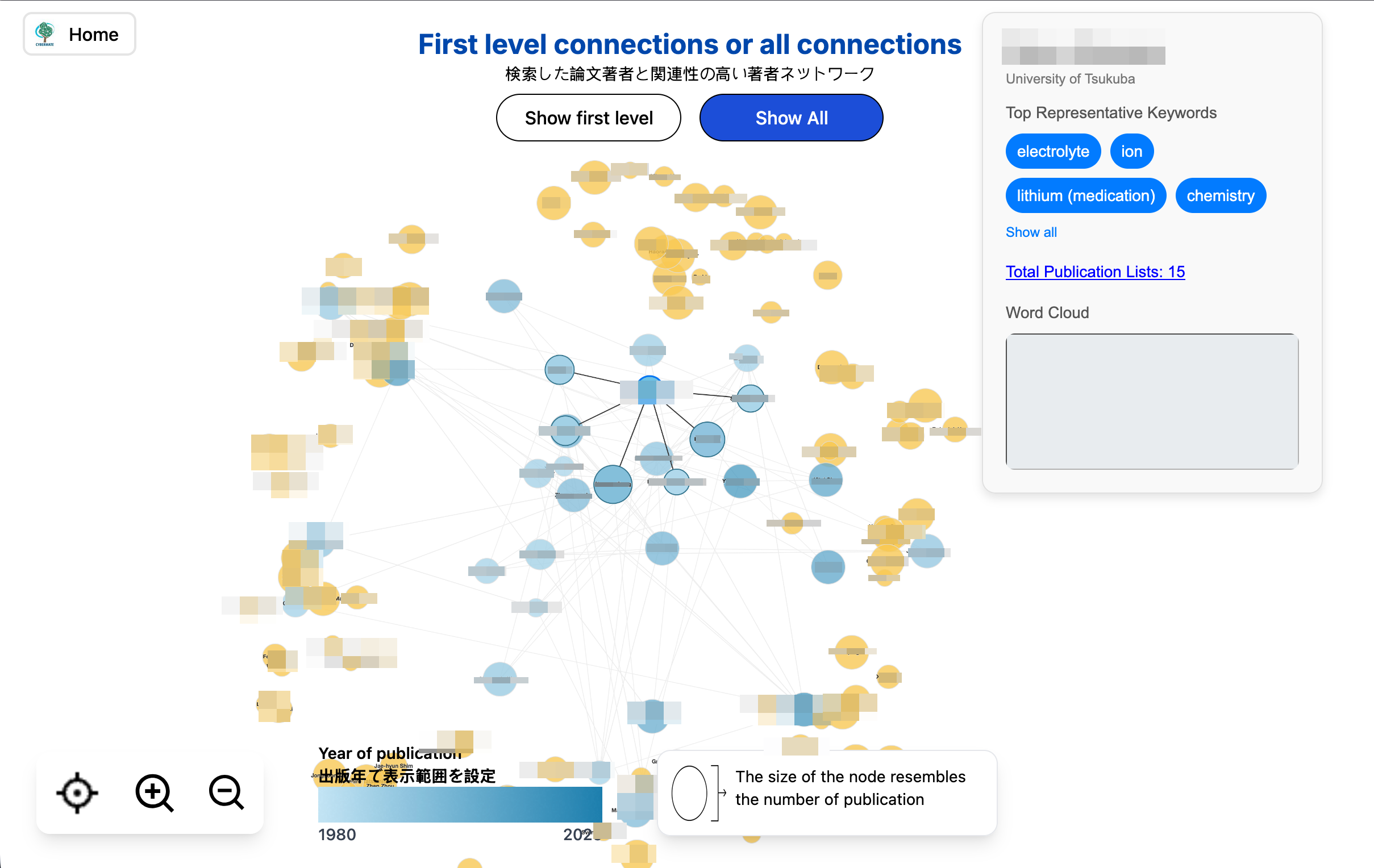}
  \caption{Example of similarity map showing all proximity co-authorship for a battery-related author. In this figure specifically, "the representative keywords" refer to the research descriptors. Author names have been omitted to preserve anonymity.}
  \label{fig:similarity-map-all-connections}
\end{figure}

In Figure~\ref{fig:similarity-map-all-connections} we provide a view with all connections, in yellow, indicating  the secondary connections, when an author is connected through a co-author to someone else. Those secondary connections, apparently not relevant, represent, in fact,  potential collaborations. 

\subsection{Serialization of the knowledge graphs}
\label{sec:rdf}

Resource Description Framework (RDF) is a standard for modelling data on the web through the triplets model subject-predicate-object~\cite{b1}. 
RDF-ready datasets can potentially be integrated together, given they adopt stable Permanent Identifiers (PID) and then explored using semantic querying through a standardized graph-based representation~\cite{doi:10.3233/SW-180294}. 

First, we enriched the dataset with Wikidata IDs corresponding to the latest author's affiliation. Wikidata~\cite{b3} is an open knowledge base providing a rich set of metadata that can be interlinked to other open data sets. 
The Wikidata knowledge base complemented the information we collected from OpenAlex as it provides stable PIDs~\cite{van2019wikidata, 10.1093/database/baw015} and more comprehensive institutional metadata that improve affiliation disambiguation and facilitate linking our records to other open datasets.

\begin{figure}[!htbp]
  \centering
  \includegraphics[width=0.45\linewidth, page=1]{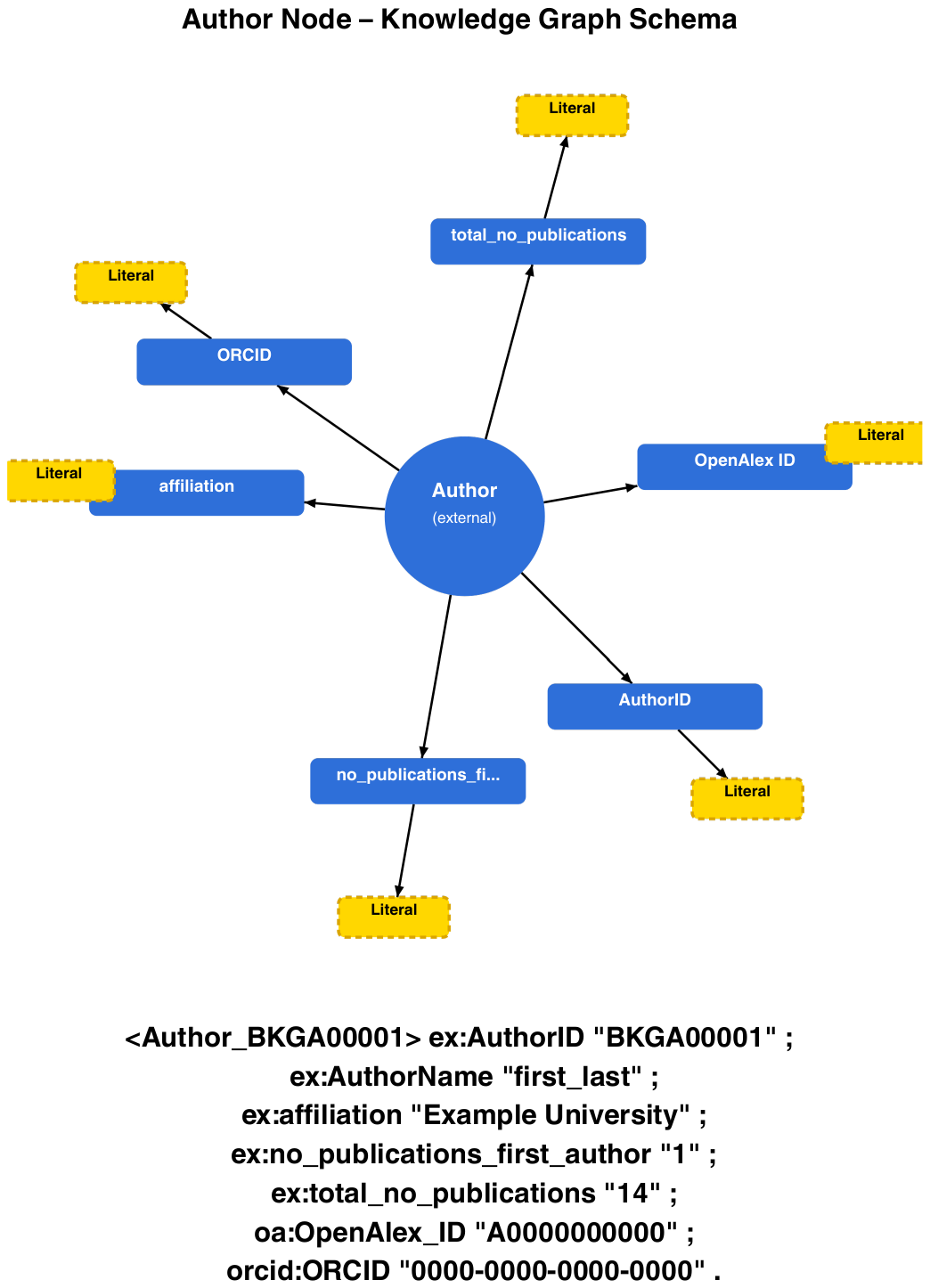}
  \includegraphics[width=0.45\linewidth, page=2]{figures/Author.pdf}
  \includegraphics[width=0.45\linewidth, page=3]{figures/Author.pdf}
  \caption{Examples of RDF serialization: the author node, representing the battery-related author, the paper node and the research descriptor node. The Wikidata ID is attached as attribute to the research descriptors. These schemata are all connected.}
  \label{fig:rdf-example}
\end{figure}

Then, we generated the RDF representation of our model composed of triples relating authors, works, and research descriptors.  
For each item (author, work, and research descriptor), we assigned a Uniform Resource Identifier (URI) that uniquely identifies the resource within our graph. We then generated triples linking authors to their works (Author–Paper relationships), papers to their publishing metadata (Paper–Publisher relationships), and papers to their associated research descriptors (Paper–Research Descriptor relationships), thereby constructing a coherent RDF graph, with Institutional Affiliation being a property of the Author node and Publisher being a property of the Paper node, (Figure~\ref{fig:rdf-example}) that captures the bibliographic and institutional relationships in a structured representation that can be queried and linked to external linked open data sources.

\section{Conclusion}
\label{sec:conclusions}

This paper presents a pipeline for constructing an author-centric battery research knowledge graph from a large open catalogue (OpenAlex) subset and for deriving interpretable author profiles from it. 
We apply our method to publications of battery research which is a domain of growing importance for the green transition. 
The principal novelty of this work lies in its author-centric framing: rather than organizing knowledge around documents or ontologies, we construct weighted research descriptor vectors that aggregate each author's thematic trajectory across time and authorship position. 
We further combine coarse-grained OpenAlex concepts with fine-grained domain keyphrases extracted from titles and abstracts, yielding richer author representations than either source alone. 
The resulting graph is serialized in RDF and linked to Wikidata, making it reusable and extensible beyond battery research.

In the future, we plan to use open-weight LLMs to increase transparency and ensure reproducibility, since more advanced models have been developed since we designed our pipeline. A stricter merging of semantically overlapping keyphrases — such as ``anode'' and ``anode material'' — could further improve word cloud readability (Figure~\ref{fig:wordcloud-a5002212606}).Additionally, further weighting factors could be considered, such as time of first publication, to reduce the penalization of younger authors or authors with a shorter publication history.

\bibliography{references}

\backmatter

\section*{Statements and Declarations}

\subsection*{Conflict of interests}

The authors declare no conflict of interests.

\subsection*{Author contributions}
\textbf{L.F.}: Conceptualisation; Methodology; Software; Data Curation; Formal Analysis; Writing -- Original Draft; 
Writing -- Review \& Editing.
\textbf{S.D.}: Conceptualisation; Methodology; Validation; Writing -- Review \& Editing; Supervision; Project Administration.
\textbf{M.Z.}: Methodology; Validation; Writing -- Review \& Editing.
\textbf{K.Ka.}: Resources; Writing -- Review \& Editing.
\textbf{K.So.}: Resources; Validation; Writing -- Review \& Editing.
\textbf{M.T.}: Visualisation; Resources; Writing -- Review \& Editing; Supervision.

\subsection*{Funding}

The authors declare that no funds, grants, or other support were received during the preparation of this manuscript.
L.F. work was done partially while at the National Institute for Materials Science until April 2024, and partially while at ScienciaLAB, as the current affiliation. 

\subsection*{Code availability}

The code is available on GitHUB at the repository \url{https://github.com/lfoppiano/cocoa-visualisation}.

\end{document}